\documentclass[runningheads]{llncs}
\usepackage{graphicx}
\usepackage{booktabs} 
\usepackage[numbers]{natbib}

\usepackage{amsmath}

\begin{document}
\title{Generalized Regression with Conditional GANs}
%
%
\author{Deddy Jobson\inst{1}
\and
Eddy Hudson\inst{2}
}
\authorrunning{Jobson and Hudson}
%
\institute{Mercari Inc., Tokyo, Japan \\
\email{deddy@mercari.com}\\
\and
University of Texas Austin, Austin, USA\\
\email{eddyhudson@utexas.edu}}
\maketitle              
\begin{abstract}
Regression is typically treated as a curve-fitting process where the goal is to fit a prediction function to data. With the help of conditional generative adversarial networks, we propose to solve this age-old problem differently; we aim to learn a prediction function whose outputs, when paired with the corresponding inputs, are indistinguishable from feature-label pairs in the training dataset. We show that this approach to regression makes fewer assumptions on the distribution of the data we are fitting to and, therefore, has better representation capabilities. We draw parallels with generalized linear models in statistics and show how our proposal extends them to neural networks. We demonstrate the superiority of this new approach to standard regression with experiments on multiple synthetic and publicly available real-world datasets, finding encouraging results, especially with real-world heavy-tailed regression datasets. To make our work more reproducible, we release our source code\footnote{Link to our code: https://github.com/deddyjobson/regressGAN}.

\end{abstract}
\section{Introduction}

Generative Adversarial Networks \citep{goodfellow_generative_2014} (GANs) revolutionized how we generate realistic artificial images. Their success is owed to the fact that compared to other methods, they can more effectively represent intractable distributions such as images. This is because instead of using hand-designed closed-form loss functions to optimize the image generator, they use an adversarial discriminator to train the generator to produce realistic images. It stands to reason that this idea can potentially be applied to better represent probability distributions in general, not just that of images. This line of reasoning is exemplified by recent advances in reinforcement learning that capitalize on the generative adversarial framework to learn a behavior policy that is hard to specify \citep{ho2016generative} \citep{hudson2022skeletal}.


We study the application of conditional GANs (CGAN) on the problem of regression with tabular data. CGANs offer an alternative approach to the training of neural networks for regression tasks. Instead of directly regressing on the target variable with a loss function like MSE, we instead train two models simultaneously, one to make predictions given only the input covariates and another to decide whether or not the predictions are distinguishable from the ground truth labels, again given the input covariates. 

The following are our contributions:
\begin{itemize}
    \item We show that using GANs for regression will require fewer assumptions on the distribution of data (Section \ref{sec:our_method}). 
   \item  We perform experiments to demonstrate the superiority of regression with GANs against other regression methods (Section \ref{sec:experiments}). 
   \item We also empirically demonstrate that training GANs for the case of regression with tabular data requires fewer tricks than with image data (Section \ref{subsec:ease_training_regressgan}).
\end{itemize}

\section{Background}

\subsection{Generative Adversarial Networks}

GANs are an unsupervised method used to learn a generative model of a probability distribution. 
Conditional GANs \citep{mirza_conditional_2014} (CGANs) were developed as a GAN-based method to generate images conditioned on the input labels. Mirza et al. used CGANs in their seminal publication to generate realistic images of numbers from the MNIST dataset based on their ground truth labels.

\subsection{Generalized Linear Models}

In regression, the objective is to maximize the likelihood of the data fitting the model. The simplest form of regression is linear regression, which assumes that the distribution of the regression residuals follows a normal distribution. 
To account for violations of the assumption of normal residuals, statisticians use the link function to extend the representation ability of linear models. This results in generalized linear models \citep{nelder_generalized_1972}.


For different industrial applications, specific link functions have been proposed. For example, to model customer revenue in e-commerce, assuming the residuals to follow heavy-tailed distributions like zero-inflated log-normal distribution \citep{wang_deep_2019}, Tweedie distribution \citep{yang_personalized_2022}, etc. have been applied with success. 
While the above methods succeed in their respective domains, none of the likelihood functions used can be generally applied to all regression problems. To be able to do that effectively, we need a likelihood function that can itself adapt to new domains.

\section{Our Method}
\label{sec:our_method}

We propose using CGANs to solve the regression problem. CGANs take a different approach to regression. Instead of maximizing the likelihood of the generated predictions belonging to the true underlying distribution of the target variable, the goal is to generate predictions as indistinguishable from the ground truths as possible. 
The benefit of this approach is that, unlike with generalized linear models, we do not need to formulate the likelihood function explicitly; with enough data, the likelihood function, too, is learned by the CGAN. 
This has been proved by Goodfellow et al., and we use the notation and derivation from their paper \citep{goodfellow_generative_2014} for our theoretical argument. More specifically, consider Proposition 1 and Equation 2 from their paper. The optimal discriminator for a fixed generator is
\begin{align}
    D^*_G(x) = \frac{p_{data}(x)}{p_{data}(x) + p_g(x)}
\end{align}
Also, from Equation 6 in Theorem 1, we see that given the optimal discriminator, the generator would optimize the Jensen Shannon divergence between the output distribution of the generator and that of the true distribution. 
\begin{align}
    C(G) = -log(4) + 2 JSD(p_{data}||p_g)
\end{align}

We see from the above equation that as long as the neural networks used in the GAN have sufficient representation capacity, we can directly optimize the Jensen Shannon Divergence between the prediction ($p_g(x)$) and true distributions ($p_{data}(x)$) without explicitly defining the true distribution.  

In this paper, we use a CGAN to represent the distribution of the target variable ($Y$) conditioned on the dependent variables ($x$). Since we use the CGAN for regression, we shall refer to our approach as the RegressGAN method. 


\section{Experiments}
\label{sec:experiments}

In order to assess the capability of CGANs for regression, we compare their performance against baseline algorithms on several datasets. 
\subsection{Datasets}

We perform experiments on three synthetic datasets and three real-world datasets. Two real-world datasets are publicly available, while the third dataset is proprietary. 

We adopt the notation commonly used in statistics. We index each data point with $i$. The values of the independent/predictor variables for each data point $i$ are represented by the vector $x_i$ and the dependent, response, or target variable by $y_i$. Finally, we represent the predictions made by any model for each data point by $\hat{y_i}$ and the residuals (errors) of predictions by $\epsilon_i$.

For all synthetic datasets, we take 100,000 random samples of data and split the data into the train (60\%), validation (20\%), and test (20\%).  

\subsubsection{Synthetic-Normal}
We synthesize a linear dataset with Gaussian noise in the following way:
\begin{align}
    \boldsymbol{\beta} &\sim MVN\left(0,\frac{1}{10} \boldsymbol{I}_{25}\right) \\
    y_i &= \mathbf{x_i} \boldsymbol{\beta} + \epsilon_i \\
    \epsilon_i &\sim \mathcal{N}(0,1) 
\end{align}

\subsubsection{Synthetic-Heteroscedastic}
Following the work of Aggarwal et al. \citep{aggarwal_benchmarking_2020}, we synthesize a dataset with Gaussian noise in the following way:
\begin{align}
    x_i &\sim N\left(0,1\right) \\
    z_i &\sim N\left(1,1\right) \\
    h_i &= (0.001+0.5|x_i|)\times z_i \\
    y_i &= x_i + h_i 
\end{align}

\subsubsection{Synthetic-Classification}
We synthesize a dataset with a binary target variable in the following way:
\begin{align}
    \boldsymbol{\beta} &\sim MVN\left(0,\frac{1}{2} \boldsymbol{I}_{25}\right) \\
    p_i &= Sigmoid(\mathbf{x_i} \boldsymbol{\beta}) \\
    y_i &\sim Bernoulli(p_i) 
\end{align}

\subsubsection{Synthetic-Tweedie}
The last synthetic dataset we use involves modeling the target variable with a distribution following the Tweedie distribution \citep{dunn_series_2005}. 
\begin{align}
    \boldsymbol{\beta} &\sim MVN\left(0,\frac{1}{10} \boldsymbol{I}_{25}\right) \\
    \mu_i &= e^{\mathbf{x_i} \boldsymbol{\beta}} \\
    y_i &\sim Tweedie(\mu_i,1.5,1)
\end{align}

\subsubsection{Car Insurance} The first real dataset we use is the French Motor Third-Party Liability Claims dataset \citep{noll_case_2020}. The dataset contains car insurance claims made over a year. We sampled 100k data points, from which we sampled 20k each for validation and test sets. 
We observe the data to have a heavy tail, as is usually the case with insurance claim data \citep{mikosch_heavy-tailed_1997}. The dataset is, therefore, appropriate for testing the representative capability of regressGAN. 


\subsubsection{Health Insurance} The next real-world dataset we use is the US Health Insurance dataset\footnote{https://www.kaggle.com/datasets/teertha/ushealthinsurancedataset}. There are 1338 records in total. We again split 20\% of the data for each of the validation and test sets. Furthermore, we use one hot encoding on the categorical features to get 8 independent variables in total, which we will use to predict the same target as before: the amount of expense claimed. 


\subsubsection{E-commerce} Lastly, we also perform experiments on the user logs provided by an online C2C marketplace where individuals can buy or sell items. We sample 100k users for our experiments. We use each user's historical data to create 12 features to forecast the future revenue generated by each user. 
Here, too, we get a zero-inflated heavy-tailed distribution for the target variable. 


\subsection{Models}
\label{subsec:models}
We compare in total three models for our experiments:
\begin{itemize}
    \item \textbf{RegressGAN}: This is our proposed method in which we use a conditional GAN for regression, which we described earlier.
    \item \textbf{FNN-MSE}: For our regression baseline, we use a feed-forward network (FNN) of the same architecture and other hyperparameters as the generator (without the noise input) with the Mean-Squared Error (MSE) loss function.
    \item \textbf{GP}: The final baseline we compare our method with is Gaussian Process Regression \citep{wang_intuitive_2022}. We use a Python implementation from scikit-learn \citep{pedregosa_scikit-learn_2011} with the RBF kernel. 
\end{itemize}



\section{Results}

To evaluate the performance of regressGAN, we measured the MAE of predictions on the test dataset since it is commonly used for evaluation when the response variable is heavy-tailed like in most of our experiments \citep{jasek_comparative_2019,glady_modified_2009}.
\begin{align}
    MAE = \frac{1}{N}\sum_{i=1}^{N}\left|\frac{y_i - \hat{y_i}}{y_i}\right|
\end{align}

We first discuss the results obtained on the synthetic datasets in Table \ref{synthetic-data-table}. Surprisingly, the regressGAN approach performs best in all but one dataset, even for the "Normal" dataset. We suspect this to be the case because while other models are prone to overfitting, the generator in regressGAN has the more difficult task of estimating the whole conditional distribution, making it more resilient to overfitting. For the heteroscedastic dataset, our results disagree with those of Aggarwal et al. in that the MSE model performs better. Perhaps it is because the underlying signal is very simple (being linear). 


\begin{table}[h]
  \caption{MAE of predictions on synthetic datasets (lower the better). Best results are highlighted in bold.} \label{synthetic-data-table}
  \begin{center}
  \begin{tabular}{cccc}
  \hline 
  \textbf{DATASET}  &\textbf{FNN-MSE}  &\textbf{GP}  &\textbf{RegressGAN (ours)} \\
  \hline 
  Normal         &\textbf{0.818}   &0.844 &\textbf{0.818} \\
  Heteroscedastic         &\textbf{0.329}   &0.851 &0.350 \\
  Classification         &\textbf{0.262}   &0.364 &\textbf{0.262} \\
  Tweedie             &0.835   &2.909  &\textbf{0.805}  \\
  \hline  
  \end{tabular}
  \end{center}
\end{table}

We next discuss the results obtained from the real-world datasets tabulated in Table \ref{real-data-table}. Unlike in the case of the synthetic datasets, the improvement in performance from regressGAN is considerable. For all datasets, we find RegressGAN to perform best among all algorithms. We suspect that the zero-inflation and high skewness of the target distributions made it difficult for the other algorithms to represent the complex distribution while respecting their strict assumptions efficiently. 

\begin{table}[]
  \caption{MAE of predictions on real datasets. Best results are highlighted in bold.} \label{real-data-table}
  \begin{center}
    \begin{tabular}{cccc}
    \hline 
    \textbf{Dataset}  &\textbf{FNN-MSE}  &\textbf{GP}  &\textbf{RegressGAN (ours)} \\
    \hline 
    Car Insurance         &0.358   &0.420 &\textbf{0.261} \\
    Health Insurance    &0.223     &0.637 &\textbf{0.178} \\
    E-commerce    &0.067     &0.093 &\textbf{0.059} \\
    \hline  
    \end{tabular}
    \end{center}
\end{table}


\section{Ablation study on tricks used to speed up GAN training}
\label{subsec:ease_training_regressgan}
We investigate tricks commonly used to help traditional GANs converge and see if we could do without them for RegressGAN. The authors who proposed GANs \citep{goodfellow_generative_2014} made changes to the objective function of the GAN to help it converge. Specifically, rather than minimizing the log probability of correct predictions by the discriminator, they maximize the log probability of incorrect predictions by the discriminator. While this subtle trick was necessary in the context of image data to help GAN training, we them unnecessary with tabular data. Our experiments found no significant difference in convergence rate or in the final MAE by forgoing the training trick. This result is encouraging because they may make RegressGANs easier to deploy in production systems. 


\section{Related Work}
Ours is not the first work to study the application of GANs for regression, though existing work is limited. Aggarwal et al. \citep{aggarwal_benchmarking_2020} considered the application of GANs to small regression problems where the distribution of the noise variable could not be explicitly modeled. 
We replicate their experiments on a synthetic heteroscedastic dataset that claims that feed-forward neural networks perform better. 
Oskarsson et al. \citep{oskarsson_probabilistic_2020} used CGANs for probabilistic regression. Rather than just learning a point estimate for each data point, the author used CGANs to estimate distributions. This work was mostly restricted to comparing different CGAN algorithms against each other.

Hudson et al. \citep{hudson2022abc} employ CGANs to regress towards action distributions in reinforcement learning that have multiple modes. In their framework, the actions are generated conditioned on the state that the agent find itself in.

A key difference between the above works and ours is in the real-world datasets chosen for the experiments; all our datasets have heavy-tailed response variables, specifically where we see the superior performance of CGANs. To our knowledge, our work is the first to discover real-world evidence of the superiority of CGANs for heavy-tailed regression.

\section{Limitations}

In this paper, we demonstrate the superiority of the GAN formulation over other loss functions, such as the MSE loss function for neural networks. However, gradient-boosted machines (GBMs) are currently the state-of-the-art method for tabular data. While neural network architectures have been proposed to surpass them, none have gained widespread adoption \citep{borisov_deep_2022} so far. Our method complements these attempts to create better neural networks for tabular data. When neural networks finally surpass GBMs over tabular data, our method can be applied to set neural networks further apart from traditional methods. 


Furthermore, recent research with extreme value theory has suggested that GANs cannot generate heavy-tailed distributions \citep{oriol_theoretical_2021,huster_pareto_2021}. This runs counter to what we have empirically observed with insurance datasets. Perhaps while RegressGAN can represent a wide variety of conditional distributions better than neural regression models, there are some distributions that it cannot represent well. 

\section{Conclusion}

In this paper, we highlight the representation capabilities of CGANs for regression tasks and show that they are superior to standard neural networks trained with the MSE loss function. They prove to be a flexible generalization of regression for neural networks in the same way generalized linear models are a flexible generalization of ordinary linear regression. 
While Oriol et al. \citep{oriol_theoretical_2021} prove theoretically that GANs cannot represent some heavy-tailed distributions, we show that GANs can better represent heavy-tailed distributions in practice than traditional methods. 
Together with our discovery of the relative ease of training GANs for tabular data, our results suggest that CGANs have the potential to be used in even more novel applications in science and industry in the years to come, especially when handling data of unknown distribution. 


\bibliographystyle{splncs04}
\bibliography{references}

\end{document}